\documentclass[conference]{IEEEtran}
\IEEEoverridecommandlockouts
\usepackage{cite}
\usepackage{amsmath,amssymb,amsfonts}
\usepackage{graphicx}
\usepackage{textcomp}

\usepackage{algpseudocode}
\usepackage{amsmath,amssymb}
\usepackage{todonotes}
\usepackage{url}
\usepackage{hyperref}

\usepackage{booktabs}
\usepackage{tabularx}
\usepackage{multirow}
\usepackage[table]{xcolor} 
\usepackage{orcidlink}

\usepackage{subcaption}
\usepackage{dblfloatfix}   

\usepackage{tikz}
\usetikzlibrary{shapes, arrows, positioning, calc, fit, backgrounds}

\def\BibTeX{{\rm B\kern-.05em{\sc i\kern-.025em b}\kern-.08em
    T\kern-.1667em\lower.7ex\hbox{E}\kern-.125emX}}
\begin{document}

\title{ECG-IMN: Interpretable Mesomorphic Neural Networks for 12-Lead Electrocardiogram Interpretation\\}

\author{
    \IEEEauthorblockN{
        Vajira Thambawita \orcidlink{0000-0001-6026-0929}\IEEEauthorrefmark{1},
        Jonas L. Isaksen \orcidlink{0000-0003-3227-1131}\IEEEauthorrefmark{2},
        Jørgen K. Kanters \orcidlink{0000-0002-3267-4910}\IEEEauthorrefmark{2}, \\
        Hugo L. Hammer \orcidlink{0000-0001-9429-7148}\IEEEauthorrefmark{3},
        Pål Halvorsen \orcidlink{0000-0003-2073-7029}\IEEEauthorrefmark{1}
    }
    \IEEEauthorblockA{
        \IEEEauthorrefmark{1} SimulaMet, Oslo, Norway\\
        \IEEEauthorrefmark{2}University of Copenhagen, Copenhagen, Denmark \\
        \IEEEauthorrefmark{3}Oslo Metropolitan University, Oslo, Norway 
        \vspace{-5pt}
    }
}

\maketitle

\begin{abstract}
Deep learning has achieved expert-level performance in automated electrocardiogram (ECG) diagnosis, yet the ``black-box'' nature of these models hinders their clinical deployment. Trust in medical AI requires not just high accuracy but also transparency regarding the specific physiological features driving predictions. Existing explainability methods for ECGs typically rely on post-hoc approximations (e.g., Grad-CAM and SHAP), which can be unstable, computationally expensive, and unfaithful to the model's actual decision-making process. In this work, we propose the ECG-IMN, an Interpretable Mesomorphic Neural Network tailored for high-resolution 12-lead ECG classification. Unlike standard classifiers, the ECG-IMN functions as a hypernetwork: a deep convolutional backbone generates the parameters of a strictly linear model specific to each input sample. This architecture enforces intrinsic interpretability, as the decision logic is mathematically transparent and the generated weights $\mathbf{W}$ serve as exact, high-resolution feature attribution maps. We introduce a transition decoder that effectively maps latent features to sample-wise weights, enabling precise localization of pathological evidence (e.g., ST-elevation, T-wave inversion) in both time and lead dimensions. We evaluate our approach on the PTB-XL dataset for classification tasks, demonstrating that the ECG-IMN achieves competitive predictive performance (AUROC comparable to black-box baselines) while providing faithful, instance-specific explanations. By explicitly decoupling parameter generation from prediction execution, our framework bridges the gap between deep learning capability and clinical trustworthiness, offering a principled path toward ``white-box'' cardiac diagnostics.
\end{abstract}


\section{Introduction}

Automated interpretation of electrocardiograms (ECGs) is fundamental to the screening and diagnosis of a wide range of cardiac pathologies. While deep learning models have demonstrated robust performance~\cite{liu2021deep} in this domain, their opaque ``black-box'' nature remains a critical obstacle to clinical adoption, regulatory approval, and trust among medical professionals~\cite{al2023explainable}.

Recent advances in interpretable deep learning~\cite{alvarez2018towards} have shown that it is possible to design models that are both accurate and explainable by construction. In particular, interpretable mesomorphic neural networks (IMNs) have demonstrated that deep hypernetworks can generate instance-specific linear models, achieving local interpretability without sacrificing global predictive performance~\cite{kadra2024interpretable}. However, existing work has primarily validated this approach on tabular data, lacking the specific architectural adaptations required for high-dimensional, temporal physiological signals.

In contrast, ECGs are highly structured, multi-lead time-series data, where clinical reasoning depends on preserving local waveform morphology (e.g., QRS complexes and ST segments) across temporal contexts. Existing explainability approaches for ECG models typically rely on post-hoc attribution methods, such as saliency maps~\cite{storaas2025evaluating} or gradient-based techniques~\cite{hicks2021explaining}, which approximate the decision boundary and often suffer from instability, limited faithfulness, and misalignment with clinical intuition.

In this work, we extend the mesomorphic interpretability paradigm~\cite{kadra2024interpretable} to high-resolution ECG time-series classification.

\medskip
\noindent
\textbf{Main contributions:}

\begin{itemize}
    \item We introduce a \textbf{Interpretable Mesomorphic Neural Network (IMN)} for ECG classification, incorporating a convolutional hypernetwork and a transition decoder that generate instance-specific, sample-wise weight maps ($\mathbf{W} \in \mathbb{R}^{12 \times L}$) directly from the input signal, where $L$ is the lenght of leads.
    \item We formulate the diagnostic decision as a strictly \textbf{local linear equation} ($y = \mathbf{W} \cdot \mathbf{X} + b$), ensuring that the generated feature attributions are faithful by design rather than post-hoc approximations.
    \item We demonstrate that the proposed model achieves \textbf{competitive performance} on the PTB-XL dataset for binary tasks (e.g., MI vs.\ Normal), comparable to standard black-box baselines.
    \item We define a clinically meaningful visualization strategy based on \textbf{aggregated temporal segments}, allowing clinicians to inspect the exact ``evidence'' used by the model for each lead and time window.
    \item We release the complete \textbf{open-source implementation} of the proposed framework, including training and evaluation code, pre-trained model checkpoints, and an interactive HuggingFace Space\footnote{\url{https://github.com/vlbthambawita/mesomorphic\_ecg}} \footnote{Check the GitHub link for the HuggingFace links}, to support \textbf{reproducible research} and transparent benchmarking.
\end{itemize}

Together, our results show that mesomorphic neural networks can be successfully adapted to structured physiological signals, offering a principled path toward transparent and trustworthy AI systems for cardiac diagnostics.

\section{Methodology: Interpretable Mesomorphic Neural Networks}

\begin{figure*}[t]
    \centering
    \pgfdeclarelayer{background}
    \pgfsetlayers{background,main}

    \begin{tikzpicture}[
        scale=0.80, 
        transform shape,
        node distance=0.5cm,
        >=stealth,
        layer/.style={
            rectangle, 
            draw=black!80, 
            fill=blue!10, 
            align=center, 
            rounded corners=2pt, 
            minimum height=1.6cm, 
            minimum width=1.4cm,
            font=\scriptsize\sffamily,
            inner sep=3pt
        },
        container/.style={
            rectangle,
            draw=black!40,
            fill=black!5,
            dashed,
            inner sep=6pt,
            rounded corners=5pt
        },
        tensor/.style={
            rectangle, 
            draw=black!60, 
            fill=white, 
            align=center,
            minimum height=1.0cm, 
            minimum width=1.0cm, 
            dashed,
            font=\scriptsize
        },
        op/.style={
            circle, 
            draw=black!80, 
            fill=orange!20, 
            align=center,
            inner sep=1.5pt, 
            font=\small\bfseries
        },
        line/.style={
            draw, 
            ->, 
            thick, 
            color=black!70
        }
    ]


    \node[tensor, label=below:{\textbf{Input} $\mathbf{X}$}] (input) {
        $12 \times L$
    };

    \node[layer, right=1.2cm of input] (b_l1) {
        \textbf{Conv1} \\[-2pt]
        \tiny $1\!\to\!16$ \\[-2pt]
        \tiny $k(3,15)$ \\[-2pt]
        \tiny BN, GELU
    };
    \node[layer, right=0.3cm of b_l1] (b_l2) {
        \textbf{Conv2} \\[-2pt]
        \tiny $16\!\to\!32$ \\[-2pt]
        \tiny $k(3,15)$ \\[-2pt]
        \tiny MaxPool \\[-2pt]
        \tiny $k(1,2)$
    };
    \node[layer, right=0.3cm of b_l2] (b_l3) {
        \textbf{Conv3} \\[-2pt]
        \tiny $32\!\to\!64$ \\[-2pt]
        \tiny $k(3,15)$ \\[-2pt]
        \tiny MaxPool \\[-2pt]
        \tiny $k(1,2)$
    };

    \node[tensor, right=0.8cm of b_l3, label=above:{\textbf{Latent} $\mathbf{Z}$}] (latent) {
        $64 \times \frac{L}{4}$
    };

    \node[layer, fill=green!10, above right=0.8cm and 1.2cm of latent] (t_l1) {
        \textbf{Stage 1} \\[-2pt]
        \tiny $64\!\to\!32$ \\[-2pt]
        \tiny $k(3,3)$ \\[-2pt]
        \tiny BN, GELU \\[-2pt]
        \tiny Up $(1,2)$
    };
    \node[layer, fill=green!10, right=0.3cm of t_l1] (t_l2) {
        \textbf{Stage 2} \\[-2pt]
        \tiny $32\!\to\!16$ \\[-2pt]
        \tiny $k(3,3)$ \\[-2pt]
        \tiny BN, GELU \\[-2pt]
        \tiny Up $(1,2)$
    };
    \node[layer, fill=green!10, right=0.3cm of t_l2] (t_l3) {
        \textbf{Proj} \\[-2pt]
        \tiny $16\!\to\!K$ \\[-2pt]
        \tiny $k(3,3)$ \\[-2pt]
        \tiny Linear
    };

    \node[tensor, right=0.8cm of t_l3, label=above:{\textbf{Gen. Weights} $\mathbf{W}$}] (weights) {
        $K \times 12 \times L$
    };

    \node[layer, fill=red!10, below right=1.2cm and 1.2cm of latent, minimum width=2.5cm] (bias_internal) {
        \textbf{AvgPool} $(1,1)$ \\[-2pt]
        $\downarrow$ \\[-2pt]
        \textbf{Linear} $64\!\to\!K$
    };

    \node[tensor, right=0.8cm of bias_internal, label=below:{\textbf{Gen. Bias} $b$}] (bias) {
        $K \times 1$
    };

    \node[op, right=0.8cm of weights, label=below:{Multiply}] (mult) {$\odot$};
    \node[op, right=0.5cm of mult, label=above:{Sum}] (sum) {$\Sigma$};
    \node[op, below=1.2cm of sum] (add) {$+$};
    \node[tensor, right=0.8cm of add, fill=yellow!20, solid, label=above:{\textbf{Logits} $y$}] (output) {
        $K$
    };

    \begin{pgfonlayer}{background}
        \node[container, fit=(b_l1) (b_l3), label=above:{\textbf{Backbone Encoder} $f_\theta$}] (backbone_box) {};
        \node[container, fit=(t_l1) (t_l3), label=above:{\textbf{Transition Decoder} $g_\phi$}] (trans_box) {};
        \node[container, fit=(bias_internal), label=above:{\textbf{Bias Gen} $h_\psi$}] (bias_box) {};
    \end{pgfonlayer}

    \draw[line] (input) -- (b_l1);
    \draw[line] (b_l1) -- (b_l2);
    \draw[line] (b_l2) -- (b_l3);
    \draw[line] (b_l3) -- (latent);

    \draw[line] (latent.east) -- ++(0.4,0) |- (t_l1.west);
    \draw[line] (latent.east) -- ++(0.4,0) |- (bias_internal.west);

    \draw[line] (t_l1) -- (t_l2);
    \draw[line] (t_l2) -- (t_l3);
    \draw[line] (t_l3) -- (weights);

    \draw[line] (bias_internal) -- (bias);

    \draw[line] (input.north) -- ++(0, 3.5) -| (mult.north) 
        node[pos=0.25, above] {\footnotesize Original Input $\mathbf{X}$ (Skip)};

    \draw[line] (weights) -- (mult);
    \draw[line] (mult) -- (sum);
    \draw[line] (sum) -- (add);
    \draw[line] (bias.east) -| (add.south)
        node[midway, right, font=\footnotesize, xshift=2pt] {Add Bias};
    \draw[line] (add) -- (output);
    
    \path let \p1=(input.west), \p2=(latent.east) in 
        node[
            anchor=north west, 
            draw=black!30, 
            fill=white, 
            dashed,
            inner sep=6pt, 
            rounded corners,
            minimum width=\x2-\x1, 
            text width=\x2-\x1-12pt,
            align=left
        ] (note) at (input.west |- backbone_box.south) [yshift=-2pt] {
            \textbf{Model Variants:}
            \begin{itemize}
                \item \textbf{Categorical:} $K$ classes ($\mathbf{W} \in \mathbb{R}^{K \times 12 \times L}$).
                \item \textbf{Binary:} Scalar score ($\mathbf{W} \in \mathbb{R}^{1 \times 12 \times L}$).
            \end{itemize}
        };

    \end{tikzpicture}
    \caption{\textbf{Detailed Architecture of the Interpretable Mesomorphic Neural Network (IMN).} 
The model operates as a hypernetwork where a deep neural network generates the parameters of a local linear model.
\textbf{(1) Parameter Generation Pathway:} The input ECG signal $\mathbf{X} \in \mathbb{R}^{12 \times L}$ is processed by a convolutional backbone encoder $f_\theta$, consisting of three stages of convolutions with asymmetric kernels $k(3,15)$ and max-pooling, to produce a compressed latent representation $\mathbf{Z}$. This latent code branches into two generators: the \textit{Transition Decoder} $g_\phi$, which utilizes upsampling and convolutions to generate high-resolution weight maps $\mathbf{W}$, and the \textit{Bias Generator} $h_\psi$, which computes scalar biases $b$ via global pooling.
\textbf{(2) Inference Pathway:} The final class logits $y$ are computed via a strictly interpretable linear equation $y = \sum(\mathbf{W} \odot \mathbf{X}) + b$, where the generated weights are applied element-wise to the original input via a skip connection.
\textbf{Notation:} $L$: input signal length; $K$: number of output classes/tasks; $k(h,w)$: convolution kernel size (height $\times$ width); $C_{in} \to C_{out}$: channel dimensions; BN: Batch Normalization; Up: Nearest-neighbor upsampling; $\odot$: Element-wise multiplication; $\Sigma$: Summation over channel and time dimensions.}
    \label{fig:imn_architecture_detailed}
\vspace{-10pt}
\end{figure*}
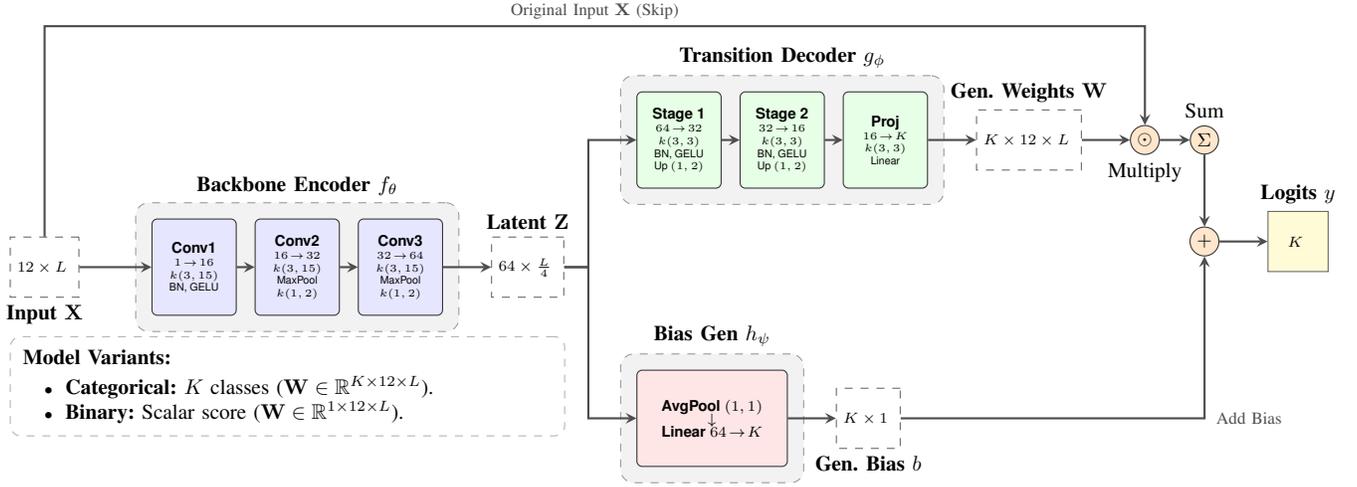

Interpretable Mesomorphic Neural Networks (IMNs) were originally proposed for tabular data, where each input feature corresponds to a semantically independent variable~\cite{kadra2024interpretable}. In this setting, a hypernetwork generates instance-specific linear weights directly over the input features, yielding faithful local explanations.

In this work, we implement an IMN specifically adapted for 12-lead ECG analysis. The IMN architecture functions as a hypernetwork: rather than directly predicting the class label, a Deep Neural Network (DNN) backbone generates the parameters (weights and biases) of a linear model specific to the input sample. This strictly local linear structure allows for intrinsic interpretability via feature attribution while maintaining the non-linear modeling capacity of deep learning. The full pipeline is illustrated in Figure~\ref{fig:imn_architecture_detailed}.

Let $\mathbf{X} \in \mathbb{R}^{C \times L}$ denote the input ECG signal, where $C=12$ is the number of leads and $L$ is the signal length (e.g., 500 or 1000 samples). We propose two variants of the IMN: a Multi-Class (Categorical) formulation and a Single-Linear (Binary) formulation.

\subsection{Hypernetwork Backbone and Transition Module}
Both formulations share a common encoder-decoder architecture designed to map the raw signal $\mathbf{X}$ to a latent weight space.

\textbf{The Encoder (Backbone):} 
The input $\mathbf{X}$ is processed by a Convolutional Neural Network (CNN) $f_\theta$ consisting of three blocks. Each block applies 2D convolutions (kernel size $3 \times 15$), Batch Normalization, GELU activation, and Max Pooling along the temporal dimension. This yields a latent feature representation $\mathbf{Z}$:
\begin{equation}
    \mathbf{Z} = f_\theta(\mathbf{X}), \quad \mathbf{Z} \in \mathbb{R}^{64 \times C \times \frac{L}{4}}
\end{equation}

\textbf{The Transition Decoder:} 
To generate element-wise weights corresponding to the original input resolution, we employ a Transition Network $g_\phi$. This module progressively upsamples the latent features using Nearest Neighbor upsampling followed by convolutions to reconstruct the temporal dimension from $\frac{L}{4}$ back to $L$. We denote the output tensor as $\mathcal{W}_{\theta,\phi}(\mathbf{X})$ to explicitly indicate its dependence on the input and model parameters:
\begin{equation}
    \mathcal{W}_{\theta,\phi}(\mathbf{X}) = g_\phi(\mathbf{Z}) = g_\phi(f_\theta(\mathbf{X}))
\end{equation}
For multi-class settings, we extract distinct weight matrices for each class $k$ by slicing the tensor along the first dimension. This is denoted as $\mathbf{W}_k = \mathcal{W}[k, :, :]$, where the slice corresponds to the generated weight map for the $k$-th class.

Simultaneously, a global bias generator $h_\psi$ applies global average pooling to $\mathbf{Z}$ followed by a linear projection to generate the scalar bias terms.

\subsection{Formulation A: Categorical IMN (Multi-Head)}
In the categorical formulation (used for standard classification tasks), the network generates a distinct weight matrix $\mathbf{W}_k$ and bias $b_k$ for each target class $k \in \{1, \dots, K\}$.

The Transition Network outputs a tensor $\mathcal{W} \in \mathbb{R}^{K \times C \times L}$. The prediction logit $z_k$ for class $k$ is computed via the dot product of the generated weights and the original input:
\begin{equation}
    \label{eq:categorical_imn}
    z_k = \sum_{c=1}^{C} \sum_{t=1}^{L} (\mathbf{W}_{k,c,t} \cdot \mathbf{X}_{c,t}) + b_k
\end{equation}
The final class probabilities are obtained via the Softmax function applied to the logit vector $\mathbf{z} = [z_1, \dots, z_K]$:
\begin{equation}
    y_k = P(y=k|\mathbf{X}) = \text{Softmax}(\mathbf{z})_k
\end{equation}

\subsection{Formulation B: Single-Linear IMN (Binary)}
For binary classification tasks where a singular decision boundary is preferred, we implement the Single-Linear formulation. Here, the Transition Network projects the latent features to a single channel output, generating $\mathbf{W} \in \mathbb{R}^{1 \times C \times L}$ and a scalar bias $b \in \mathbb{R}$.

The decision logit $z$ is computed as:
\begin{equation}
    \label{eq:single_linear_imn}
    z = \langle \mathbf{W}, \mathbf{X} \rangle + b = \sum_{c=1}^{C} \sum_{t=1}^{L} (\mathbf{W}_{c,t} \cdot \mathbf{X}_{c,t}) + b
\end{equation}
The probability of the positive class is given by the sigmoid activation $\hat{y} = \sigma(z)$. This formulation enforces that the model learns a single "evidence map" where positive values in $\mathbf{W}$ indicate support for the positive class (e.g., MI) and negative values indicate support for the negative class (e.g., Normal).
\vspace{-2pt}

\subsection{Regularization and Objective Function}
A key feature of the IMN is the enforcement of sparsity in the generated weights to enhance interpretability. We optimize the model parameters $\theta$ and $\phi$ using a composite loss function comprising a prediction loss and an L1-regularization term on the generated weights.

For Formulation A (Categorical), the loss $\mathcal{L}(\theta, \phi)$ is:
\begin{equation}
    \mathcal{L}(\theta, \phi) = \mathcal{L}_{CE}(\mathbf{y}, \mathbf{y}_{true}) + \lambda \frac{1}{K \cdot C \cdot L} \sum_{k,c,t} | \mathbf{W}_{k,c,t} |
\end{equation}
For Formulation B (Single-Linear), the loss is:
\begin{equation}
    \mathcal{L}(\theta, \phi) = \mathcal{L}_{BCE}(z, y_{true}) + \lambda \frac{1}{C \cdot L} \sum_{c,t} | \mathbf{W}_{c,t} |
\end{equation}
where $\mathcal{L}_{CE}$ is Cross-Entropy, $\mathcal{L}_{BCE}$ is Binary Cross-Entropy with Logits, and $\lambda$ is a hyperparameter controlling the sparsity of the explanation (set to $1e-4$ in our experiments).

\subsection{Intrinsic Interpretability}
Unlike post-hoc methods (e.g., Grad-CAM~\cite{selvaraju2017grad}) that approximate feature importance, the IMN provides intrinsic explanations. Because the model strictly adheres to the equation $z = \mathbf{W} \cdot \mathbf{X} + b$, the contribution of any specific segment of the signal is exactly quantified by the element-wise product:
\begin{equation}
    \mathbf{I}_{attr} = \mathbf{W} \odot \mathbf{X}
\end{equation}
where $\mathbf{I}_{attr} \in \mathbb{R}^{C \times L}$ represents the impact map. We visualize these impact maps to identify the specific ECG leads and temporal segments driving the diagnostic prediction.

\subsection{Visualization Methodology}

A primary advantage of the ECG-IMN is its intrinsic transparency, enabling exact feature attribution via the generated weights $\mathbf{W}$ without post-hoc approximations. We employ two visualization strategies depending on the output architecture, followed by a temporal aggregation step to enhance clinical readability.

\subsubsection{Intrinsic Attribution Strategies}

\textbf{Scalar Attribution (Single-Output):} In binary settings with a single output node, the decision collapses into one scalar equation $z = \sum (\mathbf{W} \odot \mathbf{X}) + b$. We define the \textbf{Scalar Impact Map} as $\mathbf{I}_{scalar} = \mathbf{W} \odot \mathbf{X}$. Here, positive contributions ($\mathbf{I}_{scalar} > 0$) strictly support the positive class (e.g., MI), while negative contributions support the negative class (e.g., Normal), effectively coupling the evidence for both outcomes in a zero-sum mechanism.

\textbf{Class-Specific Attribution (Multi-Output):} For categorical settings, the hypernetwork generates a distinct weight tensor $\mathcal{W} \in \mathbb{R}^{K \times C \times L}$. The decision logic consists of $K$ independent linear equations, yielding a \textbf{Class-Specific Impact Map} for any class $k$:
\begin{equation}
    \mathbf{I}^{(k)} = \mathcal{W}_k \odot \mathbf{X}
    \label{eq:class_impact}
\end{equation}
Unlike the scalar approach, this formulation decouples evidence. Positive regions in $\mathbf{I}^{(k)}$ explicitly support membership in class $k$, while negative regions refute it. This allows for independent assessment, identifying potential model confusion where features might simultaneously support distinct classes.

\subsubsection{Segment-wise Integration Filter}

To mitigate high-frequency noise in the raw impact maps, we implement a sliding window aggregation. For a window size $L_{win}$ and stride $S$, the \textbf{Net Signed Contribution} $C_{c,\tau}$ for channel $c$ and segment $\tau$ is computed as:
\begin{equation}
    C_{c,\tau} = \sum_{j=0}^{L_{win}-1} \mathbf{I}\left(c, \, \tau \cdot S + j\right)
    \label{eq:signed_agg}
\end{equation}
This summation functions as a semantic \textit{integration filter}. By tuning $L_{win}$ to match standard cardiac intervals (e.g., $L_{win}=250$ samples $\approx$ 0.5s), inconsistent weight oscillations cancel out ($C_{c,\tau} \to 0$), while sustained morphological patterns (e.g., ST-elevation or QRS complexes) yield high-magnitude contributions. This ensures the resulting heatmap highlights coherent clinical evidence rather than transient artifacts.

\subsection{Dataset and Preprocessing}

\textbf{The PTB-XL Database}
We evaluate our method on the PTB-XL dataset~\cite{wagner2020ptb}, a large-scale database of 21,837 clinical 12-lead ECG records from 18,885 patients. Each record is 10 seconds in duration and annotated with SCP-ECG statements, which are aggregated into five diagnostic superclasses: Normal (NORM), Myocardial Infarction (MI), ST/T Changes (STTC), Conduction Disturbance (CD), and Hypertrophy (HYP). No filtering was applied based on age or gender.

\textbf{Task Formulation: Strict Binary Classification}
While PTB-XL is natively a multi-label dataset, our study focuses on the interpretability of distinguishing specific pathologies from healthy controls. We formulate four distinct binary classification tasks: \textit{NORM vs. MI}, \textit{NORM vs. STTC}, \textit{NORM vs. CD}, and \textit{NORM vs. HYP}.

For a given target class $C_{target}$ (e.g., MI), we curate a specific subset $\mathcal{D}_{task}$ from the full database. Let $\mathcal{L}_i$ be the set of superclass labels for patient $i$. A patient is included in the binary dataset if and only if they satisfy the exclusive condition:
\begin{equation}
    (C_{target} \in \mathcal{L}_i) \oplus (\text{NORM} \in \mathcal{L}_i)
\end{equation}
where $\oplus$ denotes the logical XOR operation. This strictly removes samples with comorbidities (containing both the target pathology and NORM labels) or irrelevant conditions (containing neither), ensuring a clean decision boundary for interpretability analysis.

\textbf{Preprocessing and Splits}
We utilize the 12-lead signals sampled at 500 Hz, resulting in an input dimensionality of $\mathbf{X} \in \mathbb{R}^{12 \times 5000}$. To ensure patient independence, we adhere to the recommended stratified folds provided by the dataset authors: Folds 1--8 are used for training, Fold 9 for validation, and Fold 10 for testing. Input signals are normalized via per-lead Z-scoring:
\begin{equation}
    \mathbf{X}_{c,t}' = \frac{\mathbf{X}_{c,t} - \mu_c}{\sigma_c + \epsilon}
\end{equation}
where $\mu_c$ and $\sigma_c$ are the mean and standard deviation of lead $c$ for the specific sample.

\section{Results}
\begin{table*}[htbp]

\caption{\textbf{Quantitative Performance Evaluation on PTB-XL Binary Tasks.} 
We compare the proposed Interpretable Mesomorphic Networks against standard black-box CNN baselines across four diagnostic tasks: Normal vs.\ Conduction Disturbance (CD), Hypertrophy (HYP), Myocardial Infarction (MI), and ST/T Changes (STTC). 
Models are evaluated using both categorical (2-classes) and binary (single-linear) formulations at 100 Hz and 500 Hz sampling rates. 
``IMN Direct'' represents a naive ablation without the Transition Network (TransNet). 
\colorbox{green!20}{Green} cells denote the best performance per metric within each task, while \colorbox{red!20}{Red} cells denote the worst. 
Additionally, we highlight the internal comparison between IMN variants: \textbf{bold values} indicate the best performance between 2-classes vs.\ Binary formulations at 100 Hz, and \textcolor{red}{red text} indicates the best performance between the formulations at 500 Hz.
The results demonstrate that the interpretable IMN + TransNet models (particularly at 100 Hz) achieve predictive performance highly competitive with the black-box baselines (typically within $<$2\% AUROC), validating that intrinsic interpretability is achieved without significant degradation in diagnostic accuracy.}
\scriptsize 
\centering
\begin{tabular}{l l c c c c c c c}
\toprule
\textbf{Task} & \textbf{Model (Frequency)} & \textbf{Acc} & \textbf{B-Acc} & \textbf{Prec} & \textbf{Rec} & \textbf{F1} & \textbf{MCC} & \textbf{AUROC} \\ \midrule

\multirow{7}{*}{norm\_vs\_cd} 
& Baseline (100Hz) & \cellcolor{green!20}0.9265 & \cellcolor{green!20}0.9218 & 0.8753 & 0.9079 & \cellcolor{green!20}0.8913 & \cellcolor{green!20}0.8362 & \cellcolor{green!20}0.9732 \\
& Baseline (500Hz) & 0.8735 & 0.8830 & 0.7568 & \cellcolor{green!20}0.9112 & 0.8269 & 0.7363 & 0.9584 \\
& IMN Direct (500 Hz) & \cellcolor{red!20}0.5171 & \cellcolor{red!20}0.5114 & \cellcolor{red!20}0.3422 & \cellcolor{red!20}0.4945 & \cellcolor{red!20}0.4045 & \cellcolor{red!20}0.0215 & \cellcolor{red!20}0.5155 \\
& IMN + TransNet (2-classes) (100 Hz) & 0.9207 & 0.9053 & 0.8970 & 0.8596 & 0.8779 & 0.8197 & 0.9650 \\
& IMN + TransNet (2-classes) (500 Hz) & 0.8949 & \color{red}0.8916 & 0.8162 & \color{red}0.8816 & \color{red}0.8477 & \color{red}0.7690 & \color{red}0.9505 \\
& IMN + TransNet (Binary) (100 Hz) & \textbf{0.9251} & \textbf{0.9108} & \cellcolor{green!20}\textbf{0.9021} & \textbf{0.8684} & \textbf{0.8849} & \textbf{0.8297} & \textbf{0.9697} \\
& IMN + TransNet (Binary) (500 Hz) & \color{red}0.8953 & 0.8805 & \color{red}0.8459 & 0.8366 & 0.8412 & 0.7631 & 0.9494 \\ \midrule

\multirow{7}{*}{norm\_vs\_hyp} 
& Baseline (100Hz) & \cellcolor{green!20}0.9094 & \cellcolor{green!20}0.8550 & \cellcolor{green!20}0.8109 & 0.7589 & \cellcolor{green!20}0.7840 & \cellcolor{green!20}0.7274 & \cellcolor{green!20}0.9374 \\
& Baseline (500Hz) & 0.8942 & 0.8412 & 0.7605 & 0.7476 & 0.7540 & 0.6867 & 0.9128 \\
& IMN Direct (500 Hz) & \cellcolor{red!20}0.5092 & \cellcolor{red!20}0.5185 & \cellcolor{red!20}0.2292 & \cellcolor{red!20}0.5348 & \cellcolor{red!20}0.3209 & \cellcolor{red!20}0.0304 & \cellcolor{red!20}0.5177 \\
& IMN + TransNet (2-classes) (100 Hz) & \textbf{0.9012} & \textbf{0.8498} & \textbf{0.7795} & 0.7589 & \textbf{0.7691} & \textbf{0.7063} & 0.9276 \\
& IMN + TransNet (2-classes) (500 Hz) & 0.8665 & \color{red}0.8290 & 0.6683 & \color{red}0.7627 & 0.7124 & 0.6283 & \color{red}0.9030 \\
& IMN + TransNet (Binary) (100 Hz) & 0.8963 & 0.8494 & 0.7579 & \cellcolor{green!20}\textbf{0.7665} & 0.7622 & 0.6959 & \textbf{0.9320} \\
& IMN + TransNet (Binary) (500 Hz) & \color{red}0.8718 & 0.8249 & \color{red}0.6900 & 0.7420 & \color{red}0.7151 & \color{red}0.6332 & 0.9011 \\ \midrule

\multirow{7}{*}{norm\_vs\_mi} 
& Baseline (100Hz) & \cellcolor{green!20}0.9211 & \cellcolor{green!20}0.9103 & \cellcolor{green!20}0.9087 & \cellcolor{green!20}0.8706 & \cellcolor{green!20}0.8892 & \cellcolor{green!20}0.8285 & \cellcolor{green!20}0.9699 \\
& Baseline (500Hz) & 0.9082 & 0.8966 & 0.8890 & 0.8541 & 0.8712 & 0.8004 & 0.9676 \\
& IMN Direct (500 Hz) & \cellcolor{red!20}0.5099 & \cellcolor{red!20}0.5161 & \cellcolor{red!20}0.3779 & \cellcolor{red!20}0.5387 & \cellcolor{red!20}0.4442 & \cellcolor{red!20}0.0310 & \cellcolor{red!20}0.5149 \\
& IMN + TransNet (2-classes) (100 Hz) & 0.9003 & 0.8833 & \textbf{0.8956} & 0.8213 & 0.8569 & 0.7823 & 0.9591 \\
& IMN + TransNet (2-classes) (500 Hz) & \color{red}0.8933 & \color{red}0.8763 & \color{red}0.8833 & \color{red}0.8140 & \color{red}0.8472 & \color{red}0.7670 & \color{red}0.9555 \\
& IMN + TransNet (Binary) (100 Hz) & \textbf{0.9046} & \textbf{0.8912} & 0.8893 & \textbf{0.8423} & \textbf{0.8652} & \textbf{0.7921} & \textbf{0.9631} \\
& IMN + TransNet (Binary) (500 Hz) & 0.8857 & 0.8662 & 0.8790 & 0.7949 & 0.8348 & 0.7500 & 0.9507 \\ \midrule

\multirow{7}{*}{norm\_vs\_sttc} 
& Baseline (100Hz) & \cellcolor{green!20}0.9331 & \cellcolor{green!20}0.9254 & \cellcolor{green!20}0.9110 & 0.8989 & \cellcolor{green!20}0.9049 & \cellcolor{green!20}0.8534 & \cellcolor{green!20}0.9792 \\
& Baseline (500Hz) & 0.9095 & 0.9060 & 0.8565 & 0.8941 & 0.8749 & 0.8045 & 0.9685 \\
& IMN Direct (500 Hz) & \cellcolor{red!20}0.5233 & \cellcolor{red!20}0.5174 & \cellcolor{red!20}0.3708 & \cellcolor{red!20}0.4971 & \cellcolor{red!20}0.4248 & \cellcolor{red!20}0.0333 & \cellcolor{red!20}0.5105 \\
& IMN + TransNet (2-classes) (100 Hz) & \textbf{0.9233} & 0.9130 & \textbf{0.9028} & 0.8779 & 0.8902 & 0.8315 & 0.9748 \\
& IMN + TransNet (2-classes) (500 Hz) & \color{red}0.8936 & \color{red}0.8879 & \color{red}0.8372 & 0.8683 & \color{red}0.8525 & \color{red}0.7696 & \color{red}0.9589 \\
& IMN + TransNet (Binary) (100 Hz) & 0.9230 & \textbf{0.9199} & 0.8775 & \cellcolor{green!20}\textbf{0.9094} & \textbf{0.8932} & \textbf{0.8333} & \textbf{0.9757} \\
& IMN + TransNet (Binary) (500 Hz) & 0.8706 & 0.8783 & 0.7701 & \color{red}0.9046 & 0.8319 & 0.7341 & 0.9543 \\ \bottomrule

\end{tabular}
\label{tbl_results}
\vspace{-10pt}
\end{table*}

\begin{figure*}[t]
    \centering

    \begin{subfigure}[t]{0.32\textwidth}
        \centering
        \includegraphics[page=1,width=\linewidth, trim=80 150 60 65,clip]{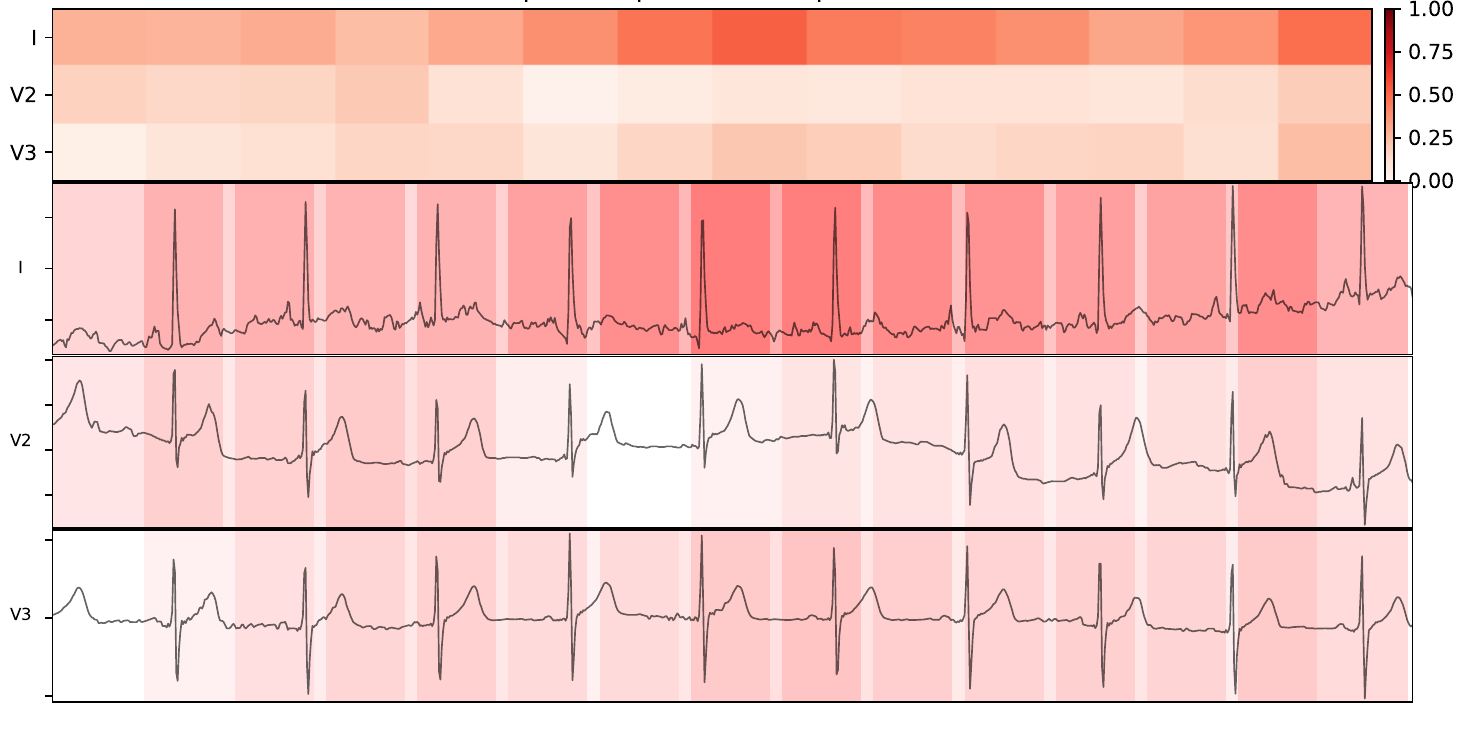}
        \caption{\textbf{Grad-CAM} ($L_{win}=125$, $S=67$)}
        \label{fig:res100_a}
    \end{subfigure}
    \hfill
    \begin{subfigure}[t]{0.32\textwidth}
        \centering
        \includegraphics[page=1,width=\linewidth, trim=80 150 60 65,clip]{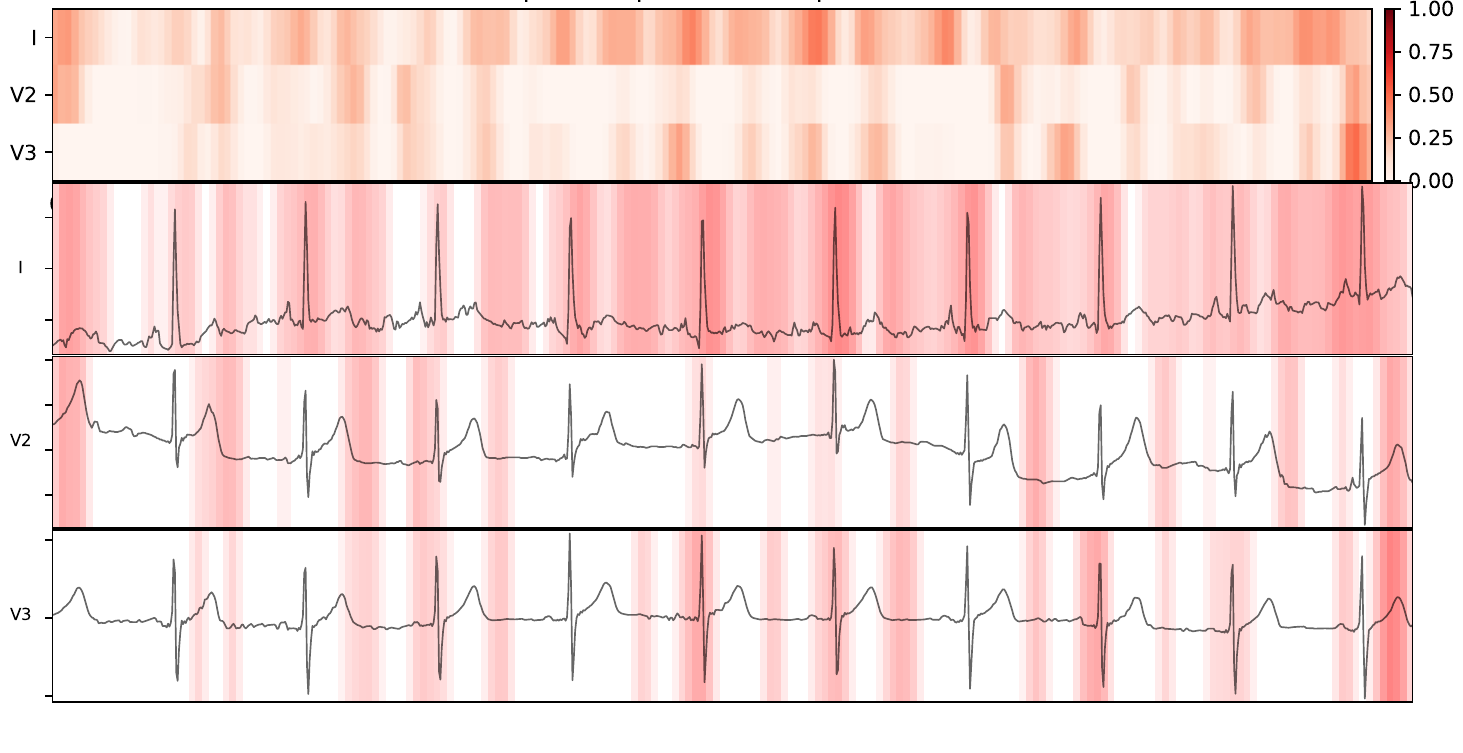}
        \caption{\textbf{Grad-CAM} ($L_{win}=10$, $S=5$)}
        \label{fig:res100_b}
    \end{subfigure}
    \hfill
    \begin{subfigure}[t]{0.32\textwidth}
        \centering
        \includegraphics[page=1,width=\linewidth, trim=80 150 60 65,clip]{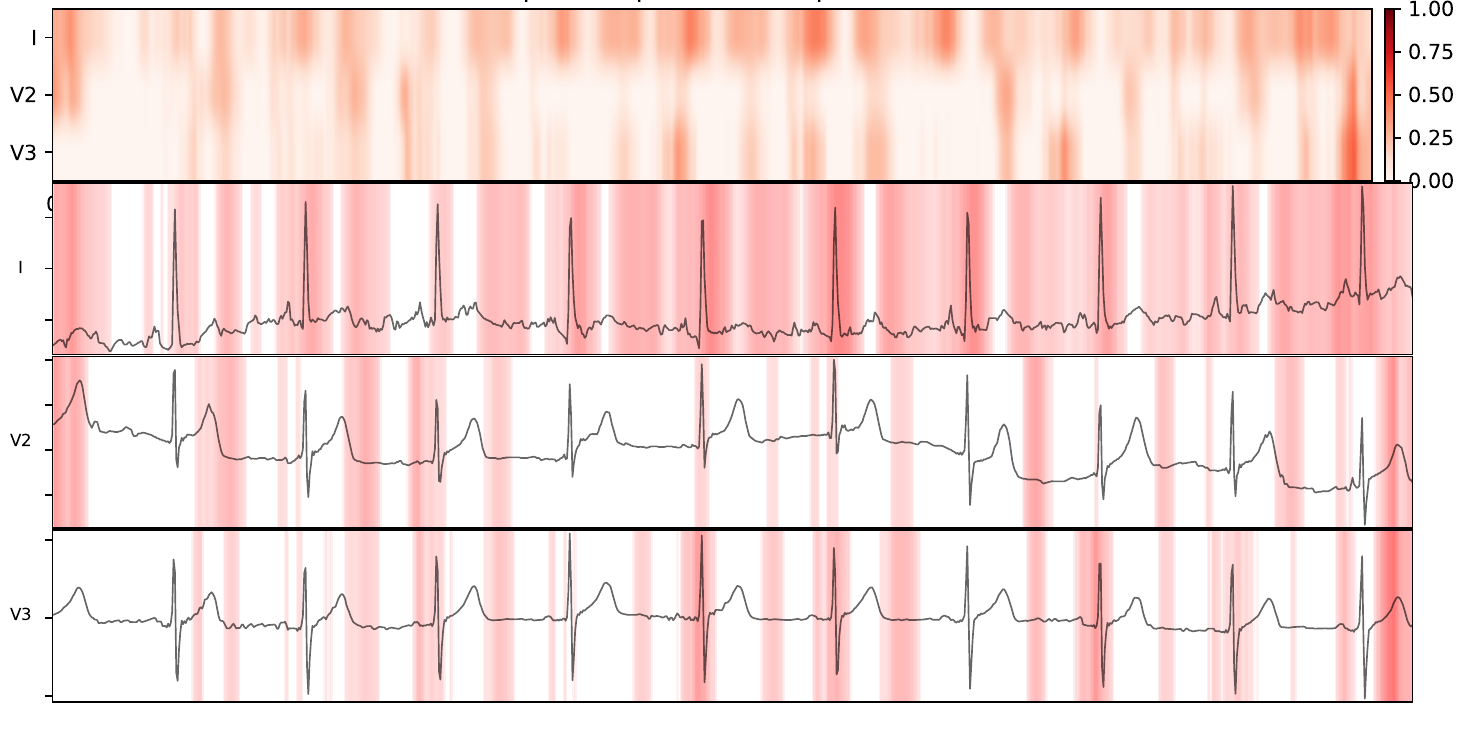}
        \caption{\textbf{Grad-CAM} ($L_{win}=2$, $S=1$)}
        \label{fig:res100_c}
    \end{subfigure}
    
     \vspace{0.6em}
    \begin{subfigure}[t]{0.32\textwidth}
        \centering
        \includegraphics[page=1,width=\linewidth, trim=0 100 0 15,clip]{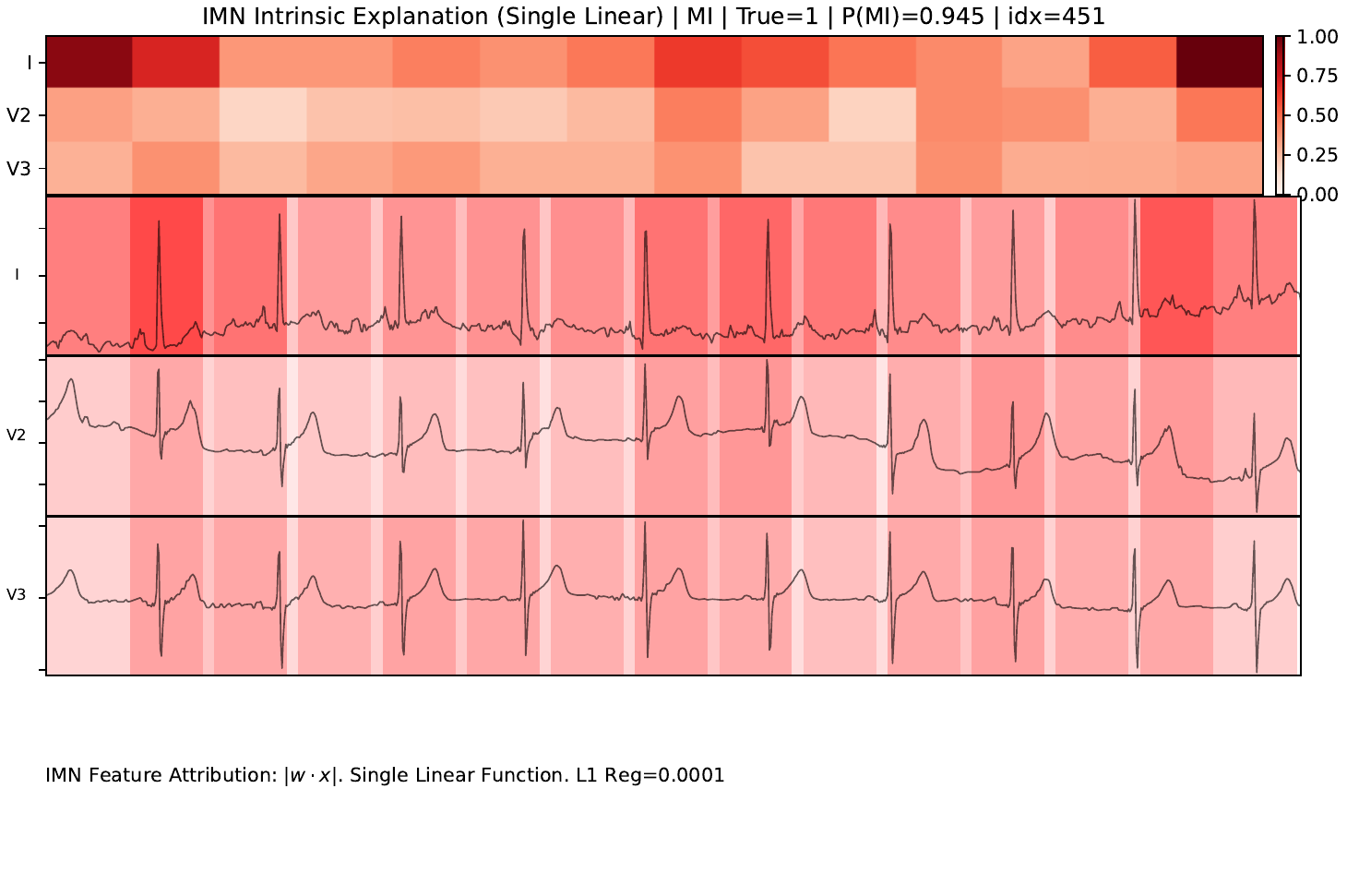}
        \caption{\textbf{Sing:-Lin: IMN} ($L_{win}=125$, $S=67$)}
        \label{fig:res100_a}
    \end{subfigure}
    \hfill
    \begin{subfigure}[t]{0.32\textwidth}
        \centering
        \includegraphics[page=1,width=\linewidth, trim=0 100 0 15,clip]{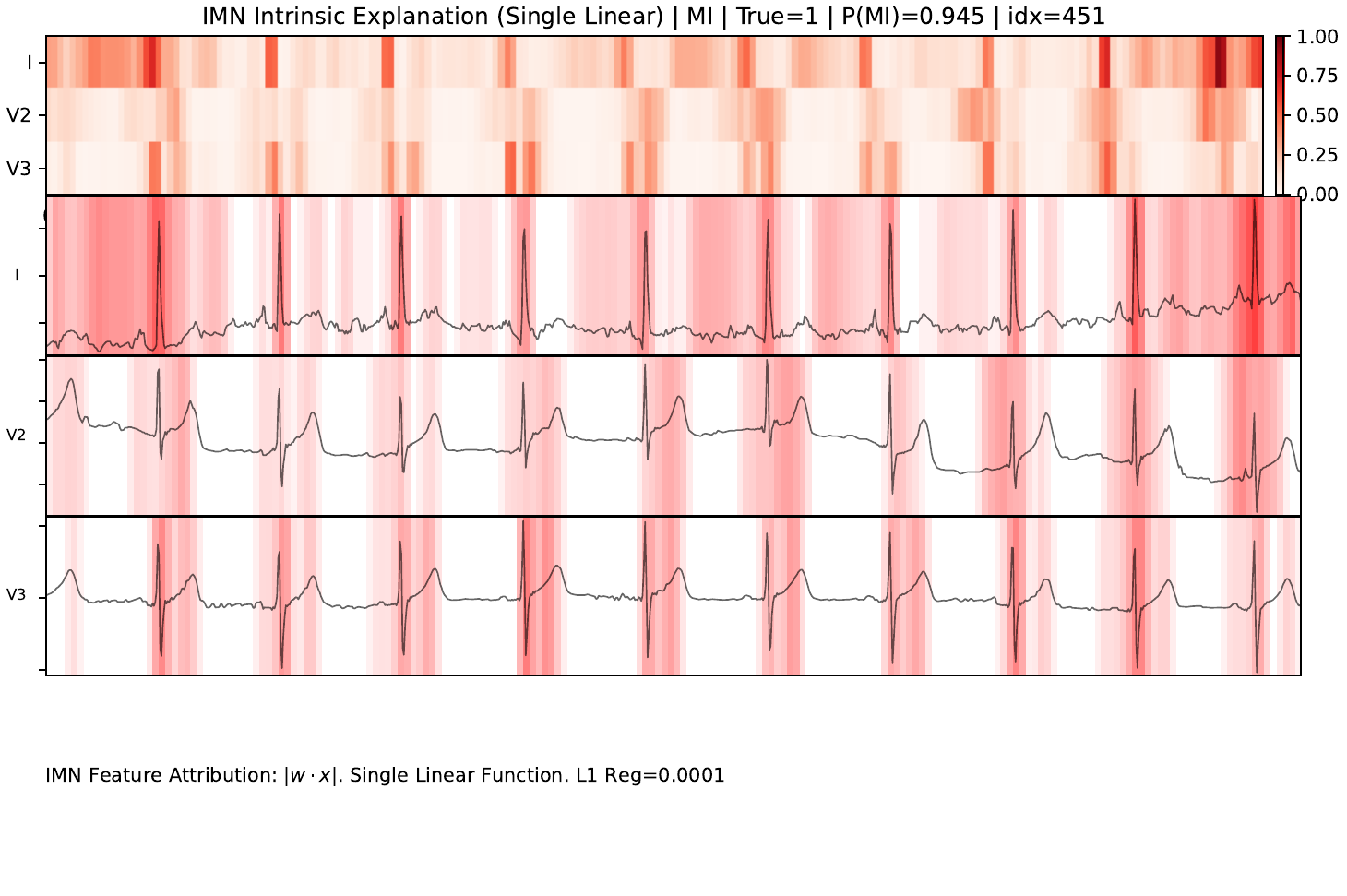}
        \caption{\textbf{Sing:-Lin: IMN} ($L_{win}=10$, $S=5$)}
        \label{fig:res100_b}
    \end{subfigure}
    \hfill
    \begin{subfigure}[t]{0.32\textwidth}
        \centering
        \includegraphics[page=1,width=\linewidth, trim=0 100 0 15,clip]{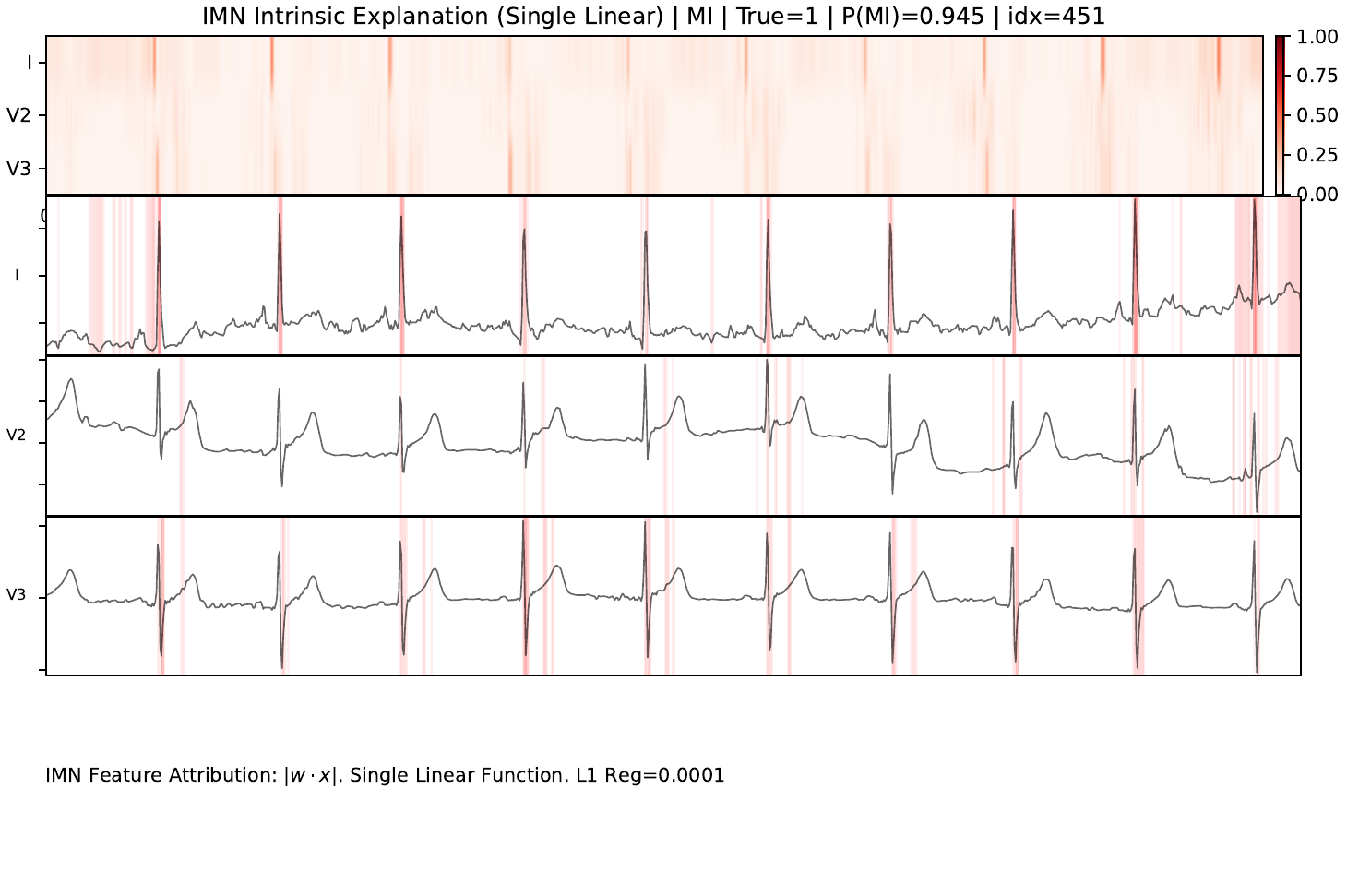}
        \caption{\textbf{Sing:-Lin: IMN} ($L_{win}=2$, $S=1$)}
        \label{fig:res100_c}
    \end{subfigure}

    \vspace{0.6em}

    \begin{subfigure}[t]{0.32\textwidth}
        \centering
        \includegraphics[page=1,width=\linewidth, trim=0 100 0 15,clip]{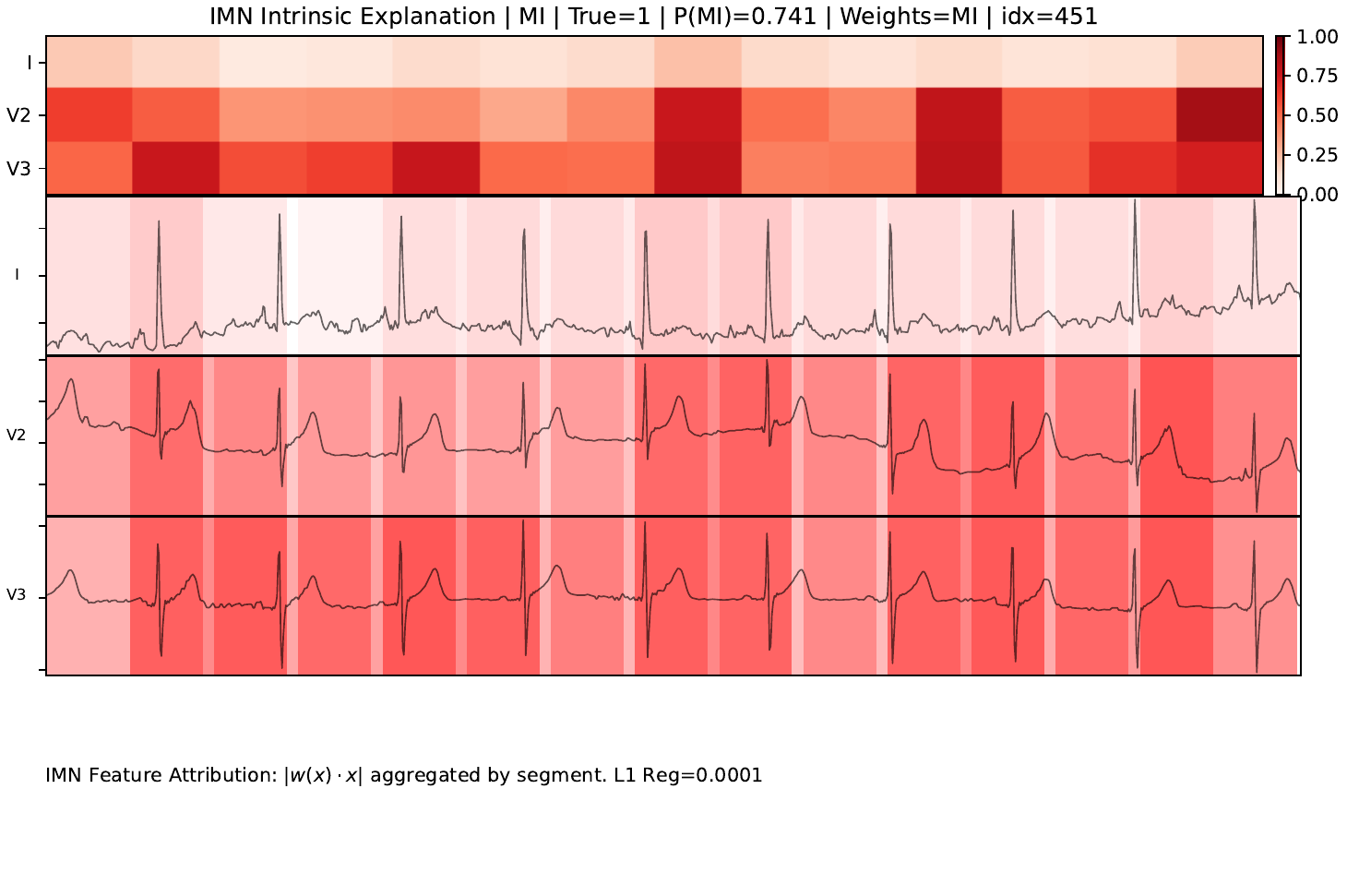}
        \caption{\textbf{Categorical IMN} ($L_{win}=125$, $S=67$)}
        \label{fig:res100_d}
    \end{subfigure}
    \hfill
    \begin{subfigure}[t]{0.32\textwidth}
        \centering
        \includegraphics[page=1,width=\linewidth, trim=0 100 0 15,clip]{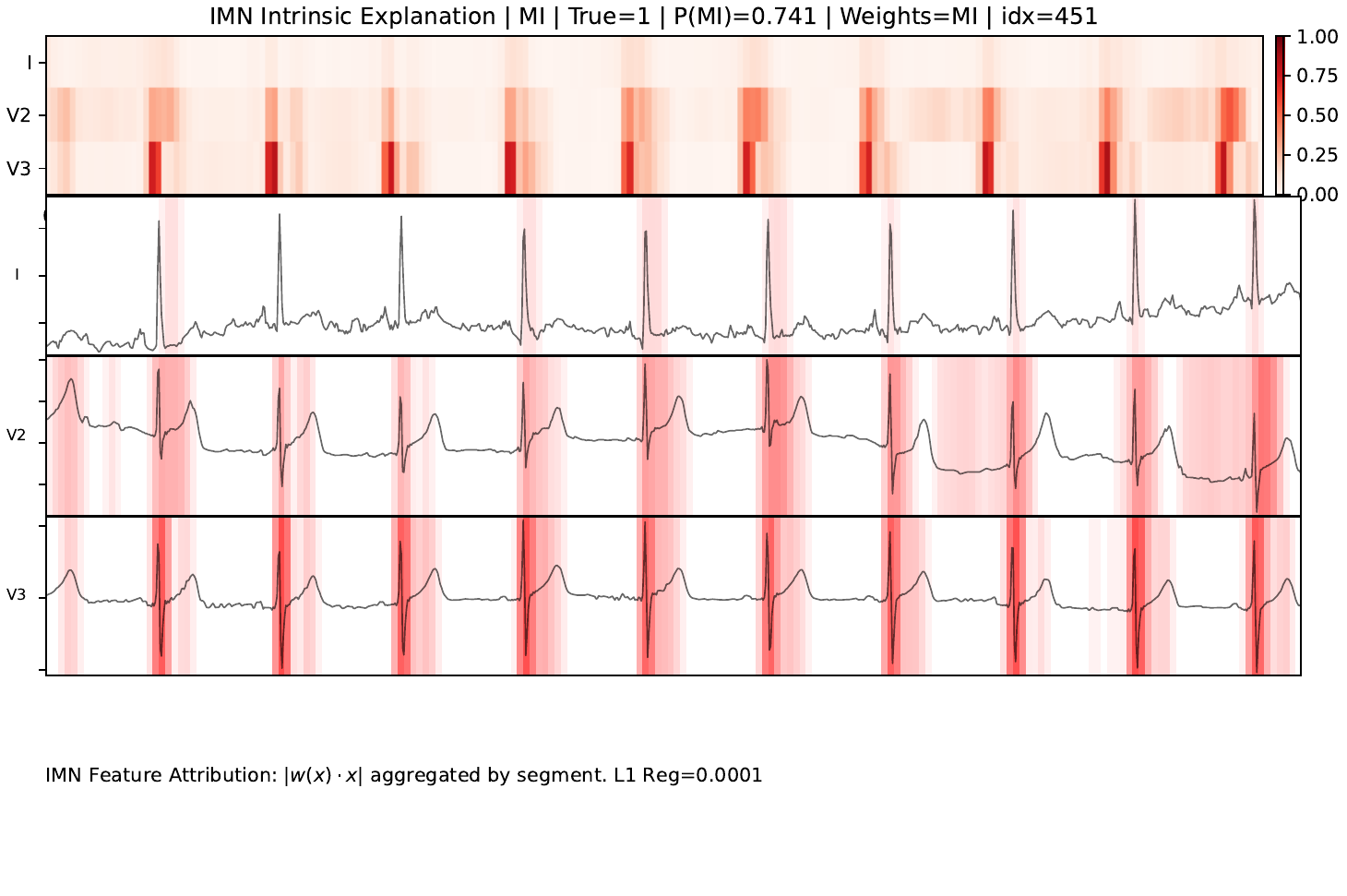}
        \caption{\textbf{Categorical IMN} ($L_{win}=10$, $S=5$)}
        \label{fig:res100_e}
    \end{subfigure}
    \hfill
    \begin{subfigure}[t]{0.32\textwidth}
        \centering
        \includegraphics[page=1,width=\linewidth, trim=0 100 0 15,clip]{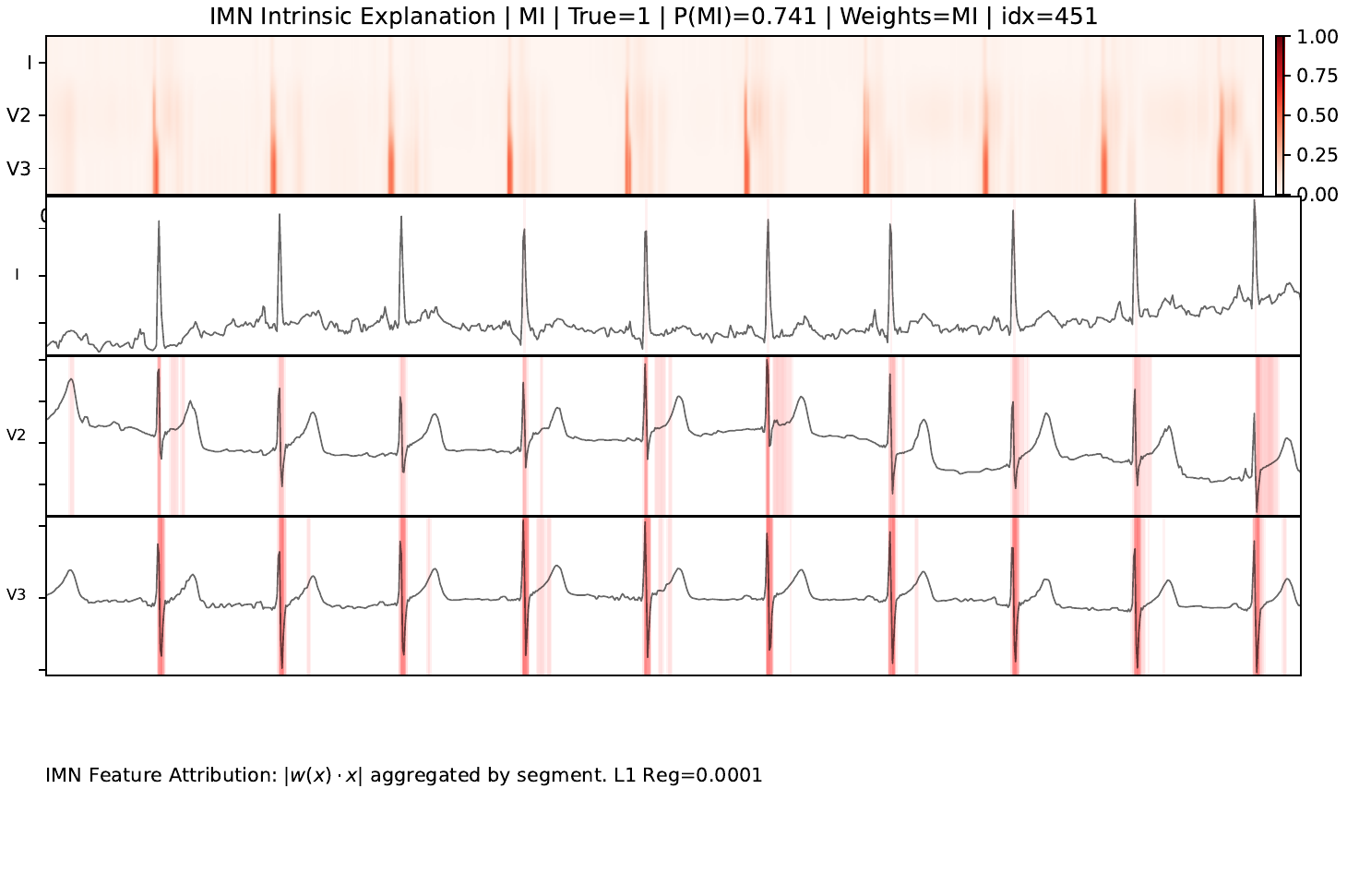}
        \caption{\textbf{Categorical IMN} ($L_{win}=2$, $S=1$)}
        \label{fig:res100_f}
    \end{subfigure}

   \caption{\textbf{Comparison of Grad-CAM and IMN-based intrinsic attribution across aggregation scales (100 Hz).}
The top row shows attribution maps obtained using a modified Grad-CAM, visualizing class-discriminative activations for a fixed window and stride. The middle row presents intrinsic importance maps from the Single-Linear IMN, and the bottom row shows those from the Categorical IMN, all evaluated using identical inputs and window--stride settings. Red intensity denotes positive contribution to the myocardial infarction (MI) prediction, with opacity proportional to contribution magnitude. For visual clarity, only three representative ECG leads are displayed. Predicted MI probabilities from the best-performing checkpoints are $P(\mathrm{MI})=0.556$ for Grad-CAM, $P(\mathrm{MI})=0.945$ for the Single-Linear IMN, and $P(\mathrm{MI})=0.741$ for the Categorical IMN.}

\label{fig:resolution_comparison_100hz}
\vspace{-10pt}
\end{figure*}

\subsection{Quantitative Performance Evaluation}
We evaluated the proposed simple IMN Network against a black-box CNN baseline on the PTB-XL dataset across four binary diagnostic tasks: Normal vs. Conduction Disturbance (CD), Hypertrophy (HYP), Myocardial Infarction (MI), and ST/T Changes (STTC). Table~\ref{tbl_results} summarizes the results across both 100 Hz and 500 Hz sampling rates.

The results demonstrate that the IMN architectures incorporating the Transition Network (TransNet) achieve predictive performance highly competitive with opaque black-box baselines, typically maintaining an AUROC gap of less than 2\%. For instance, in the \textit{norm\_vs\_mi} task at 500 Hz, the binary IMN formulation achieved an AUROC of 0.9631 , closely trailing the baseline AUROC of 0.9792. Similarly, for \textit{norm\_vs\_sttc}, the categorical IMN at 100 Hz achieved an AUROC of 0.9748.

Crucially, the ablation study denoted as "IMN Direct" (naive generation without the Transition Network) resulted in near-random performance (e.g., AUROC $\approx$ 0.51 for \textit{norm\_vs\_mi}). This confirms that the Transition Network is essential for effectively mapping latent features to high-resolution sample-wise weights.

\subsection{Impact of Aggregation Granularity}
To assess the interpretability of the generated feature maps, we analyzed the impact of window size ($L_{win}$) and stride ($S$) on the visualization of intrinsic importance ($W \odot X$). As illustrated in Fig. \ref{fig:resolution_comparison_100hz}, we compared three resolution levels using 100 Hz sampling.

\subsection{Interactive Visualization and Ablation Interface}

To enhance interpretability and facilitate clinical validation, we developed an interactive web application hosted on Hugging Face Spaces (\url{https://huggingface.co/spaces/SEARCH-IHI/mesomorphicECG_XAI}) using Gradio. The interface enables real-time exploration of IMN predictions on the PTB-XL test fold based on the single-linear or categorical formulation, where the decision logit is computed as $\text{logit} = \sum (w \cdot x) + b$.

The application provides three core functionalities. First, it visualizes instance-specific feature attribution ($w \cdot x$), with adjustable window and stride parameters to aggregate contributions and generate interpretability heatmaps over the 12-lead ECG signals. Second, it highlights the most influential contributors by marking the top-$k$ leads and temporal segments with the highest signed contributions, alongside the ground-truth clinical report for direct comparison. Third, it enables real-time counterfactual ablation by allowing users to remove selected leads or segments and recompute the predicted probability $P(\text{pos\_class})$, quantifying their impact on the model output.

This interactive framework demonstrates that IMN predictions are driven by localized, clinically meaningful ECG features. By allowing direct manipulation of input components and observing corresponding prediction changes, the tool provides practical validation of the model's intrinsic interpretability.

\section{Conclusion}

In this work, we introduced the ECG-IMN, a novel framework that adapts Interpretable Mesomorphic Neural Networks to the domain of high-resolution 12-lead ECG classification. By designing the model as a hypernetwork that generates parameters for a strictly local linear equation ($y=W\cdot X+b$), we successfully bridge the gap between the high predictive capacity of deep learning and the rigorous transparency required for clinical trustworthiness. Unlike prevalent post-hoc explanation methods, which can be unfaithful to the underlying model, our approach ensures that the generated feature attributions are intrinsic to the decision-making process.

Our evaluation on the PTB-XL dataset confirms that the enforceability of interpretation does not come at the cost of diagnostic accuracy. The ECG-IMN achieves predictive performance comparable to standard black-box CNN baselines, with AUROC scores typically within a $2\%$ margin for key binary tasks such as Myocardial Infarction detection. Furthermore, our ablation studies highlight the critical role of the proposed Transition Decoder in effectively mapping latent representations to sample-wise weight maps, enabling the precise localization of pathological evidence across leads and temporal segments.

Finally, by providing clinically meaningful visualization strategies and an open-source interactive interface, we empower clinicians to validate model reasoning against medical knowledge. This work demonstrates that mesomorphic neural networks offer a viable and principled direction for developing ``white-box'' diagnostic systems that are both accurate and inherently transparent.

\section{Future Work}

While the IMN establishes a transparent framework for ECG analysis, several avenues remain for enhancing its efficacy and clinical utility.

\subsection{Architectural and Generative Extensions}
Future research should explore alternative backbones for the hypernetwork, such as 1D-ResNets or Transformers, to better capture long-range temporal dependencies and refine the generated weight maps $\mathbf{W}$. Additionally, optimizing the Transition Network scaling laws could reduce parameter complexity without sacrificing fidelity. Beyond classification, integrating IMN-derived linear feedback into ECG generative models~\cite{thambawita2021deepfake} offers a promising direction for improving the physiological realism of synthetic data compared to black-box discriminators.

\subsection{Adaptive Visualization and Clinical Integration}
The interpretability of the generated feature maps is currently sensitive to fixed aggregation parameters. We plan to develop \textit{adaptive windowing} techniques that dynamically adjust the visualization segment ($L_{win}$) based on the patient's heart rate (R-R interval) to consistently capture morphological features like QRS complexes. To validate clinical utility, future work will quantify the alignment between IMN attribution maps and expert-annotated regions of interest. Finally, we aim to extend the bias generator to incorporate multi-modal patient metadata (e.g., age, sex), thereby refining the decision boundary for diverse patient populations.

\section*{Acknowledgment}
This work is part of the European project SEARCH, which is supported by the Innovative Health Initiative Joint Undertaking (IHI JU) under grant agreement No. 101172997. The JU receives support from the European Union’s Horizon Europe research and innovation programme and COCIR, EFPIA, Europa Bio, MedTech Europe, Vaccines Europe, Medical Values GmbH, Corsano Health BV, Syntheticus AG, Maggioli SpA, Motilent Ltd, Ubitech Ltd, Hemex Benelux, Hellenic Healthcare Group, German Oncology Center, Byte Solutions Unlimited, AdaptIT GmbH. Views and opinions expressed are however those of the author(s) only and do not necessarily reflect those of the aforementioned parties. Neither of the aforementioned parties can be held responsible for them. 

\bibliographystyle{IEEEtran}
\bibliography{references}

\end{document}